\documentclass[conference]{IEEEtran}
\IEEEoverridecommandlockouts
\usepackage{amsmath,amssymb,amsfonts}
\usepackage{algorithmic}
\usepackage{graphicx}

\usepackage{float}
\usepackage{textcomp}
\usepackage{biblatex}
\usepackage{xcolor}
\usepackage{multicol}
\usepackage{subcaption}
\usepackage{longtable}
\usepackage{lipsum}
\usepackage{booktabs, capt-of, tabularx}
\usepackage{cuted, stfloats}
\usepackage{hyperref}
\usepackage{url}
\hypersetup{
    colorlinks=true,
    linkcolor=black,
    filecolor=black,
    citecolor=black,
    urlcolor=black,
}

\begin{document}
\title{Leveraging Parameter Efficient Training Methods for Low Resource Text Classification: A Case Study in Marathi}
\makeatletter
\newcommand{\newlineauthors}{%
  \end{@IEEEauthorhalign}\hfill\mbox{}\par
  \mbox{}\hfill\begin{@IEEEauthorhalign}
}
\makeatother

\author{\IEEEauthorblockN{Pranita Deshmukh}
\IEEEauthorblockA{
\textit{Pune Institute of Computer Technology}\\
\textit{L3Cube Pune}\\
Pune, India \\
dpranita9158@gmail.com}
\and
\IEEEauthorblockN{Nikita Kulkarni}
\IEEEauthorblockA{
\textit{Pune Institute of Computer Technology}\\
\textit{L3Cube Pune}\\
Pune, India \\
nikitakulkarni0108@gmail.com}
\and
\IEEEauthorblockN{Sanhita Kulkarni}
\IEEEauthorblockA{
\textit{Pune Institute of Computer Technology}\\
\textit{L3Cube Pune}\\
Pune, India \\
sanhitak17@gmail.com}
\and
\newlineauthors
\IEEEauthorblockN{Kareena Manghani}
\IEEEauthorblockA{
\textit{Pune Institute of Computer Technology}\\
\textit{L3Cube Pune}\\
Pune, India \\
kareenamanghani@gmail.com}
\and
\IEEEauthorblockN{Raviraj Joshi}
\IEEEauthorblockA{\textit{Indian Institute of Technology Madras} \\
\textit{L3Cube Pune}\\
Pune, India \\
ravirajoshi@gmail.com}
}

\maketitle

\begin{abstract}
With the surge in digital content in low-resource languages, there is an escalating demand for advanced Natural Language Processing (NLP) techniques tailored to these languages. BERT (Bidirectional Encoder Representations from Transformers), serving as the foundational framework for numerous NLP architectures and language models, is increasingly employed for the development of low-resource NLP models. Parameter Efficient Fine-Tuning (PEFT) is a method for fine-tuning Large Language Models (LLMs) and reducing the training parameters to some extent to decrease the computational costs needed for training the model and achieve results comparable to a fully fine-tuned model.
In this work, we present a study of PEFT methods for the Indic low-resource language Marathi. We conduct a comprehensive analysis of PEFT methods applied to various monolingual and multilingual Marathi BERT models. These approaches are evaluated on prominent text classification datasets like MahaSent, MahaHate, and MahaNews. The incorporation of PEFT techniques is demonstrated to significantly expedite the training speed of the models, addressing a critical aspect of model development and deployment.
In this study, we explore Low-Rank Adaptation of Large Language Models (LoRA) and adapter methods for low-resource text classification. We show that these methods are competitive with full fine-tuning and can be used without loss in accuracy. This study contributes valuable insights into the effectiveness of Marathi BERT models, offering a foundation for the continued advancement of NLP capabilities in Marathi and similar Indic languages.

\end{abstract}

\begin{IEEEkeywords}
Natural Language Processing, Bidirectional Encoder Representations from Transformers, Parameter Efficient Fine-Tuning, Low-Rank Adaptation of Large Language Models, Adapter methods, Marathi, Datasets.
\end{IEEEkeywords}

\section{Introduction}

BERT (Bidirectional Encoder Representations from Transformers) is a machine learning-based model used for NLP (Natural Language Processing)\cite{vaswani2017attention}. It can be used to perform a variety of language tasks such as Sentiment Analysis, Question answering, Summarization, etc.  It forms the basis of various recent NLP architectures which include OpenAI’s GPT-2, XLNet, RoBERTa, etc. BERT has been specifically crafted to work on generating pre-trained bidirectional representation from unlabeled text. It achieves this by concurrently considering both left and right contextual information across all layers of the model during the pre-training process \cite{devlin2018bert}. BERT has been widely used and trained to work with English vocabulary but in recent years it has been trained and fine tuned to work on various Indic languages such as Hindi and Marathi. MahaBERT, MahaAIBERT, and MahaRoBERTa are all BERT-based language models which have been trained on the Marathi corpus \cite{joshi2022l3cube}.

Though Large Language Models (LLMs) based on transformer architecture and BERT have achieved great results as these models become larger their fine-tuning becomes infeasible on consumer hardware. In addition to this storage and deployment of these fine-tuned models for downstream tasks become expensive and time-consuming as well. Parameter-Efficient Fine-Tuning (PEFT) approaches address these issues. PEFT\footnote{\url{https://huggingface.co/blog/peft}} methods fine-tune only a selective small number of model parameters; freezing most of the parameters of the pre-trained LLMs, hence, reducing the computational power and storage costs required to train these pre-trained LLMs. Thus PEFT methods enable to achieve results considering only a limited number of trainable parameters as compared to a full fine-tuned model.

This paper aims to provide a comparative overview of mono and multilingual BERT models trained with and without PEFT methods. These models have been trained on 3 Marathi datasets which are MahaSent, MahaHate, and MahaNews, and are available on
L3Cube Github Repository\footnote{\url{https://github.com/l3cubepune/MarathiNLP}}. These models are evaluated based on their running time and their accuracy metric. 

The PEFT methods that have been used are LoRA (Low-Rank Adaptation of Large Language Models) and Adapter method \cite{hu2021lora, pfeiffer2020adapterhub}. The various mono and multilingual marathi based BERT models used are MahaBERT, MahaAIBERT, MahaRoBERTa, Muril-BERT, IndicBERT and multilingual-cased BERT.

The paper is structured as follows: Section \ref{sec:Related Works} includes information about related work in this domain. Section \ref{sec:Datasets} describes the datasets used in the study.The models that were used are described in the \ref{sec:Methodology}, along with the methodology used model training and the accuracies obtained. Section \ref{sec:Result} provides the details about the results, and Section \ref{sec:Conclusion} consists of conclusions based on the results.

\section{Related Works}\label{sec:Related Works}

Devlin.et.al \cite{devlin2018bert} introduces a new language representation model called BERT (Bidirectional Encoder Representations and Transformers)  that takes into account both left and right context in all layers. It attained satisfactory results on certain NLP tasks which include improving; GLUE score to 80.5\%, MultiNLI accuracy to 86.7\%, SQuAD v1.1 1 question answering Test F1 to 93.2 and SQuAD v2.0 Test f1 to 83.1. Grail.et.al \cite{grail-etal-2021-globalizing} presents a new hierarchical propagation layer that uses a hierarchical technique to handle input blocks individually, dispersing information across several transformer windows. In comparison to standard and pre-trained language-model-based summarizers, the proposed method is validated on three extractive summarization corpora of lengthy scientific papers and news articles, showing state-of-the-art results for long document summarization and comparable performance for smaller document summarization.

Kowsher.et.al \cite{9852438} introduces Bangla-BERT, a monolingual model specifically designed for the Bangla language, outperforms multilingual BERT and other non-contextual models in various NLP tasks, including named entity recognition and binary language classification. Pre-trained on a large 40 GB dataset, Bangla-BERT achieves superior results in Banfakenews, Sentiment Analysis on Bengali News Comments, and Cross-lingual Sentiment Analysis in Bengali, surpassing previous state-of-the-art findings by significant margins. It surpasses these findings by 3.52\%, 2.2\%, and 5.3\%, respectively. Jha.et.al \cite{9418387} introduces a multilingual IndicBERT which works on 12 indian languages and consists of 10 times fewer parameters as compared to other models. An Innovation and Entrepreneurship project at CMR Institute of Technology implemented an Agri-commerce Services search based on Question Answering, achieving a high accuracy level of 92\% using Indic BERT for diverse Indian languages like Tamil, Gujarati, Telugu, Kannada, and Malayalam. 

Kalaivani and Thenmozhi \cite{kalaivani2021multilingual} Hate speech and offensive language detection mechanism for English and various Indo-Aryan Languages. They discover various models which perform 2 basic subtasks mainly, detection of Hate speech and offensiveness in comments in Marathi, Hindi and English and classification of these comments into Profanity, Hate speech and Offensive in Hindi and English languages. They implemented the RoBERTa model for 1 task in English, BERT for the 2nd English task and MBERT for implementing these tasks in Hindi and Marathi. The macro-averaged F1-scores attained in this study   were 0.7919, 0.7320, 0.8223, 0.6242, and 0.5110,for English task 1, Hindi task 1, Marathi task 1, English task 2, and Hindi task 2 respectively.

In the study conducted by Mohta.et.al \cite{liu2022few} they have compared in-context learning (ICL) and parameter-efficient fine-tuning (PEFT) methods on various pre-trained models i.e T0 an encoder-decoder Transformer model. They concluded that PEFT achieved better accuracy with lower computational costs. The study conducted by Pavlyshenko \cite{pavlyshenko2023analysis} considered the possibility of fine-tuning the Llama GPT-2 large language model (LLM) for conducting the multitask analysis of financial news.  They utilized the PEFT/LoRA approach for fine-tuning the model. The model was fine-tuned for tasks which included: summarizing a text, analyzing a text from the perspective of the financial markets, and highlighting key aspects of a text.

\section{Datasets}\label{sec:Datasets}

\begin{table}[!h]
\begin{center}
\begin{tabular}{|p{0.03\linewidth}|p{0.25\linewidth}|p{0.08\linewidth}|p{0.08\linewidth}|p{0.07\linewidth}|p{0.07\linewidth}|p{0.08\linewidth}|}
    \hline
    \textbf{Sr. No} & \textbf{Dataset Name} & \textbf{Total size} & \textbf{Train Size} & \textbf{Test Size} & \textbf{Eval Size} & \textbf{Labels}\\
    \hline
    1 & L3CubeMahaSent Dataset & 18378 & 12114 & 2250 & 1500 & 3\\
    \hline
    2 & L3Cube-MahaHate & 25000 & 21500 & 2000 & 1500 & 4\\
    \hline
    3 & L3Cube-MahaNews & 27525 & 22014 & 2761 & 2750 & 12\\
    \hline

\end{tabular}
\caption{Dataset Train-Test-Validation Split}
\label{tab:Datasets}
\end{center}
\end{table}

For this survey, we experimented on three Marathi datasets viz. MahaSent,MahaHate and MahaNews respectively.All these datasets contain Marathi texts and their corresponding semantic labels. Following is a short jist of these datasets -\\
\subsection{L3CubeMahaSent Dataset \textsuperscript{\cite{kulkarni2021l3cubemahasent}}}
It is a sentiment dataset in marathi language consisting of tweets that are manually labeled on three classes mainly- Positive(1), Negative(-1) and Neutral(0) for sentiment analysis. This large dataset consists of 18,378 tweets out of which 15,864 tweets were divided into train (tweets-train.csv), test(tweets-test.csv) and validation(valid-test.csv) categories and the remaining 2,514 tweets are assigned in tweets-extra.csv if needed for any additional study.

\subsection{L3Cube MahaHate\textsuperscript{\cite{velankar2022l3cube}}}
L3Cube-MahaHate is a trailblazing and all-encompassing dataset consisting of manually labeled tweets in Marathi. There are two different formats of the dataset- a 4-class dataset and a 2-class dataset. For this study we have considered the 4-class dataset consists of 25,000 tweet samples in total, classified into categories like hate, offensive, profane, and not. This subset includes 1,500 validation samples, 2,000 test samples, and 21,500 training samples, which together offer a strong basis for training and assessment. 

\subsection{L3Cube-MahaNews\protect\footnote{https://github.com/l3cube-pune/MarathiNLP/tree/main/L3Cube-MahaNews}}
L3Cube-MahaNews dataset is thoughtfully divided into three different segments based on article length: long (LDC), medium (LPC), and short (SHC). The dataset is made more specific and granular by further subdividing each category into 13 classes. For this study, we have considered the LDC category consisting of articles longer than 1,000 words that provide extensive coverage and in-depth analysis. There are twelve classes in the dataset covering a wide range of subjects: Auto, Bhakti, Crime, Education, Fashion, Health, International, Manoranjan, Politics, Sports, Technology, and Travel. In addition to improving the dataset's usability, this careful categorization captures the variety of Marathi news sources. With applications ranging from topic modeling to sentiment analysis across the numerous domains represented in this extensive collection, researchers and developers using the L3Cube-MahaNews dataset are well-positioned to make notable advancements in Marathi natural language processing.

\begin{figure}[h]
    \centering
    \includegraphics[scale=0.22]{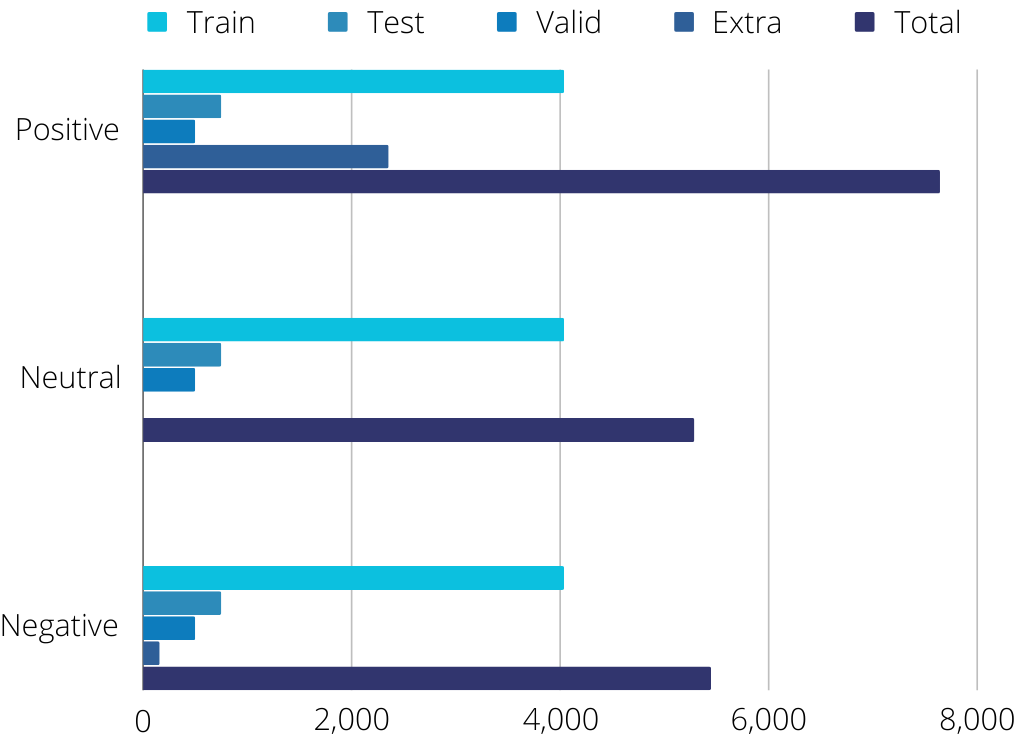}
    \caption{L3Cube MahaSent Dataset}
    \label{fig:Datasets}
\end{figure}
\begin{figure}[h]
    \centering
    \includegraphics[scale=0.22]{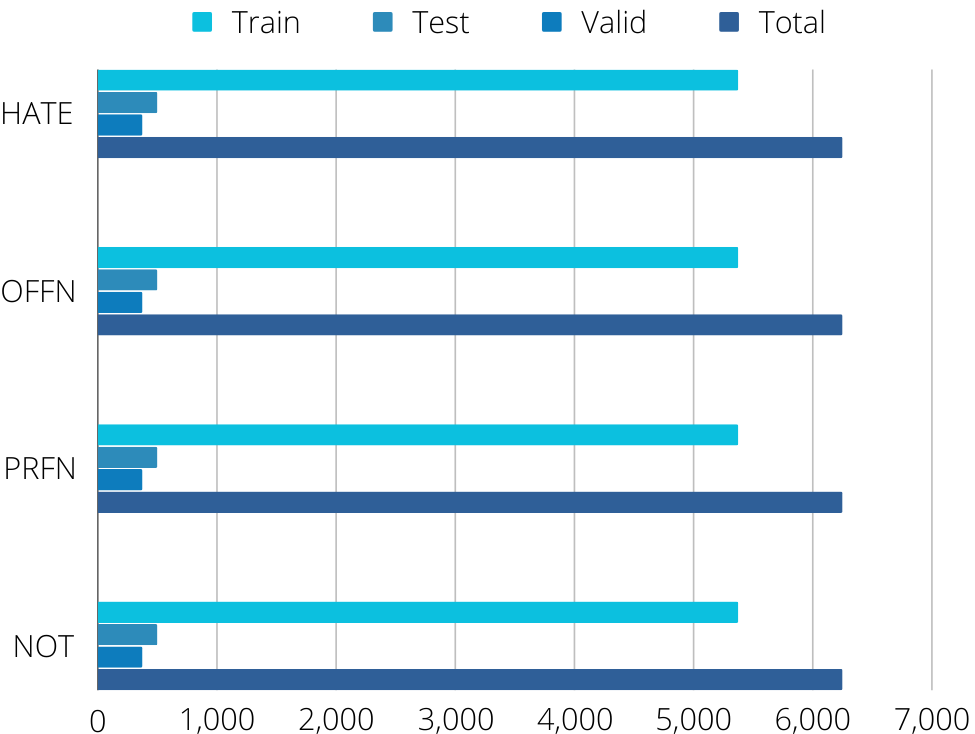}
    \caption{L3Cube MahaHate Dataset}
    \label{fig:Datasets}
\end{figure}
\begin{figure}[!h]
    \centering
    \includegraphics[scale=0.22]{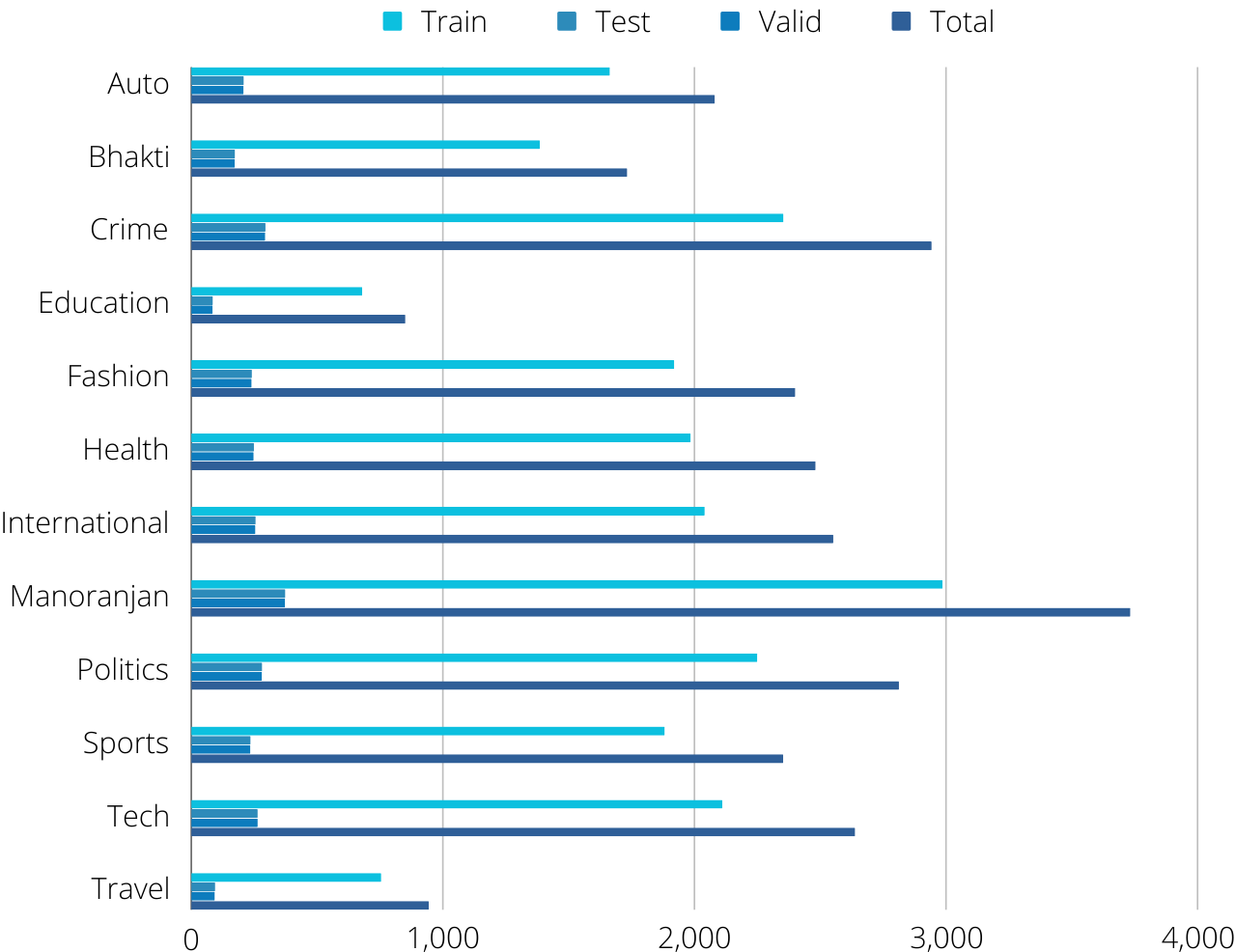}
    \caption{L3Cube MahaNews Dataset}
    \label{fig:Datasets}
\end{figure}

\newpage
\section{Methodology}\label{sec:Methodology}
The BERT based transformer models considered for this study are MahaBERT, MahaRoBERTa, multilingual base cased model, MuRIL base cased BERT model.
\begin{figure*}[h]
    \centering
    \includegraphics[scale=0.45]{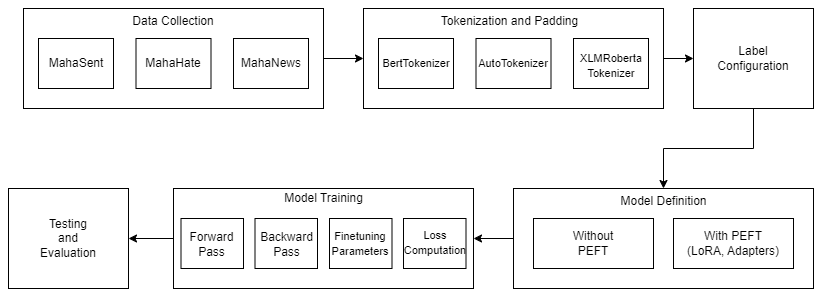}
    \caption{FlowChart}
    \label{fig:Datasets}
\end{figure*}
\subsection{Data Collection}
As mentioned above, we carefully selected three diverse datasets - MahaSent, Mahahate, and MahaNews - encompassing various domains and linguistic variations within Marathi text. This ensured our models could generalize well across different contexts.
\subsection{Tokenization and Padding}
Text needs to be broken down into smaller units for model processing. We utilized various tokenizers: BertTokenizer for its diverse vocabulary and subword segmentation, AutoTokenizer for its adaptability, and XLMRobertaTokenizer specifically designed for multilingual tasks. To ensure efficient batch processing, we padded sentences to a max length of 200 tokens.
\subsection{Model Definition}
We evaluated four distinct pre-trained models:  MahaBERT, MahaRoberta, Multilingual-cased, and Muril. Each model underwent training both with and without Parameter Efficient Fine Tuning(PEFT) techniques. The initial phase involved the conventional training of the models, while the subsequent phase incorporated PEFT to assess its impact on model efficiency and performance. PEFT selectively fine-tunes a fraction of pre-trained language model parameters, reducing costs while maintaining competitive performance.

The LoRA and Adapter mechanisms were incorporated as PEFT strategies to adaptively scale model parameters and improve training performance.
\vspace{12pt}
\subsubsection{Transformer Models}

We used the following models in our research:

\paragraph{MahaBERT\textsuperscript{\cite{joshi2022l3cube}}}
MahaBERT\footnote{\url{https://huggingface.co/l3cube-pune/marathi-bert-v2}} is a Marathi BERT model based on Google's muril-base-cased version of the basic BERT architecture. It was fine-tuned using the L3Cube-MahaCorpus and several publicly available Marathi monolingual datasets. The model was initially pretrained using next sentence prediction (NSP) and masked language modeling (MLM) on 104 languages, including Marathi.

\paragraph{MahaRoBERTa\textsuperscript{\cite{joshi2022l3cube}}}
MahaRoBERTa\footnote{\url{https://huggingface.co/l3cube-pune/marathi-roberta}} is a Marathi RoBERTa model based on XLM-roberta-base. Optimized for the Marathi monolingual dataset and L3Cube-MahaCorpus, it outperforms mBERT on various tasks. MahaRoBERTa's hyperparameter adjustments from the original BERT improve its performance. The model is pre-trained on 100 languages, including Marathi, using the MLM objective.

\paragraph{BERT-base-multilingual-cased\textsuperscript{\cite{devlin2018bert}}}
Developed by Google, the pre-trained BERT (Bidirectional Encoder Representations from Transformers) multilingual basic model (cased)\footnote{\url{https://huggingface.co/bert-base-multilingual-cased}} is designed to comprehend and provide multilingual contextualized word and phrase representations. "Cased" denotes that the model keeps track of case, allowing it to differentiate between capital and lowercase letters.

\paragraph{MuRIL-base-cased\textsuperscript{\cite{khanuja2021muril}}}
MuRIL\footnote{\url{https://huggingface.co/google/muril-base-cased}} is a BERT-based model pretrained using a range of corpora, including Wikipedia, Common Crawl, PMINDIA, and Dakshina, for 17 Indian languages. It introduces unique adaptations, such as the use of translation and transliteration segment pairings in training, in contrast to the conventional multilingual BERT technique.
\vspace{12pt}
\subsubsection{PEFT Methods}
During fine-tuning, Parameter-Efficient Fine Tuning (PEFT) methods add a few trainable parameters (the adapters) on top of the pretrained model parameters that have been frozen. The adapters are trained to pick up knowledge relevant to a given task.
\paragraph{LoRA}
Low-rank decomposition (LoRA)\footnote{\url{https://huggingface.co/docs/peft/conceptual_guides/lora}} makes fine-tuning easier by reducing trainable parameters while maintaining frozen pre-trained weights. This method supports numerous compact models and smoothly interacts with other efficiency strategies, while offering performance that is similar to fully-fine-tuned models without introducing inference latency. Transformer attention blocks are usually addressed by LoRA, whose efficacy depends on the chosen rank and the shape of the original weight matrix.
The LoRA parameters considered during this study are:
\begin{itemize}
    \item \textbf{Task Type:} Downstream task type.(task\_type=TaskType.SEQ\_CLS)
    \item \textbf{Inference Mode:} Training mode
    \item \textbf{r:} Rank of LoRA update matrices.(r\_values = [2, 5, 10, 20, 50])

    \item \textbf{lora\_alpha:} Scaling factor for LoRA updates.( lora\_alpha\_values = [8, 16, 32])

    \item \textbf{lora\_dropout:} Dropout rate for LoRA adapters.( lora\_dropout\_values = [0.1, 0.3, 0.5]
)
    \item \textbf{Bias:} Bias training strategy
\end{itemize}
\paragraph{Adapters}
Small modules known as adapters\footnote{\url{https://huggingface.co/docs/transformers/main/peft}} are inserted between layers of trained models to enhance performance on subsequent challenges while maintaining the original weights. When compared to fine-tuning, they are more effective in resource-constrained contexts because they achieve parameter efficiency, which lowers the number of trainable parameters. Adapters enable for task-specific adaptation and reduce catastrophic forgetting while preserving the model's broad knowledge.

The Adapter parameters used during this study were:
\begin{itemize}
    \item \textbf{mh\_adapter:} Allows MoE adapters to function better for multi-head attention..(mh\_adapter=False)

    \item \textbf{reduction\_factor:} A controllable adapter size that affects loss and performance.(reduction\_factor=0.25)
    \item \textbf{Non\_linearity:} Indicates the activation function for adapters.(non\_linearity="gelu")
    \item \textbf{output\_adapter:} Allows for the use of a final output adapter for potential loss reduction.(output\_adapter=True)

\end{itemize}

\subsection{Testing and Evaluation}
Each dataset had a distinct "test set" specifically reserved for performance evaluation. This ensures the model is assessed on data it hasn't encountered during training, avoiding biased results. To facilitate efficient model processing, the test set was transformed into a standardized "tensor dataset" format defined by Hugging Face. This structured representation incorporates input IDs, attention masks, and labels, offering streamlined data access and manipulation.The tensor dataset was loaded into a "dataloader". The model passed each data batch through its internal layers, generating predictions for the entity labels within each instance. The predicted labels were compared to the actual labels in the test set.
We primarily used accuracy as the metric to assess model performance. High accuracy scores signify strong model capabilities in recognizing and classifying named entities within the test set.

\begin{table}[!h]
    \centering
    \begin{tabular}{|c|c|c|c|}
        \hline
        \textbf{Models} & \textbf{Normal} & \textbf{LoRA} & \textbf{Adapter} \\
        \hline
        MahaBert-v2 & 84.9&	85.5&	83.8 \\
        \hline
        Maha-Roberta & 83.7&	84.76&	84.09 \\
        \hline
        multilingual-cased & 81.69	&81.53	&80.67 \\
        \hline
        MuRIL-based & 84.3	&84.93	&82.76 \\
        \hline
    \end{tabular}
    \caption{L3Cube MahaSent Accuracy Values.}
    \label{tab:Table II}
\end{table}
\begin{figure}[!h]
    \centering
    \includegraphics[scale=0.20]{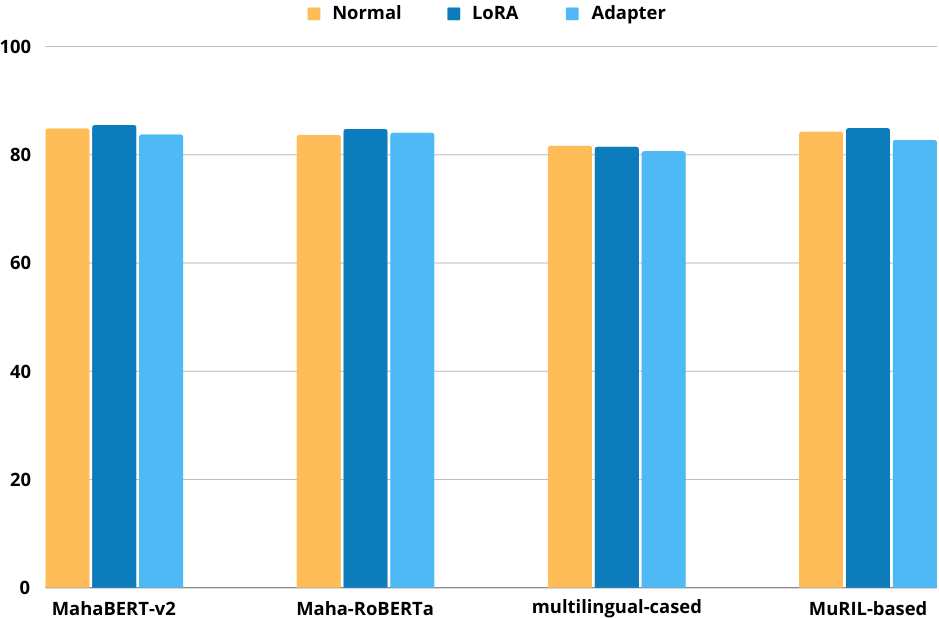}
    \caption{L3Cube MahaSent Accuracy.}
    \label{fig:Datasets}
\end{figure}
\begin{table}[!h]
    \centering
    \begin{tabular}{|c|c|c|c|}
        \hline
        \textbf{Models} & \textbf{Normal} & \textbf{LoRA} & \textbf{Adapter} \\
        \hline
        MahaBert-v2 & 81.55&	82.95	&81.8\\
        \hline
        Maha-Roberta & 80	&81.8	&82 \\
        \hline
        multilingual-cased & 78.3	&80.5	&79.78\\
        \hline
        MuRIL-based &81.2	&81.47	&79.39 \\
        \hline
    \end{tabular}
    \caption{L3Cube MahaHate Accuracy Values.}
    \label{tab:Table III}
\end{table}
\begin{figure}[h]
    \centering
    \includegraphics[scale=0.20]{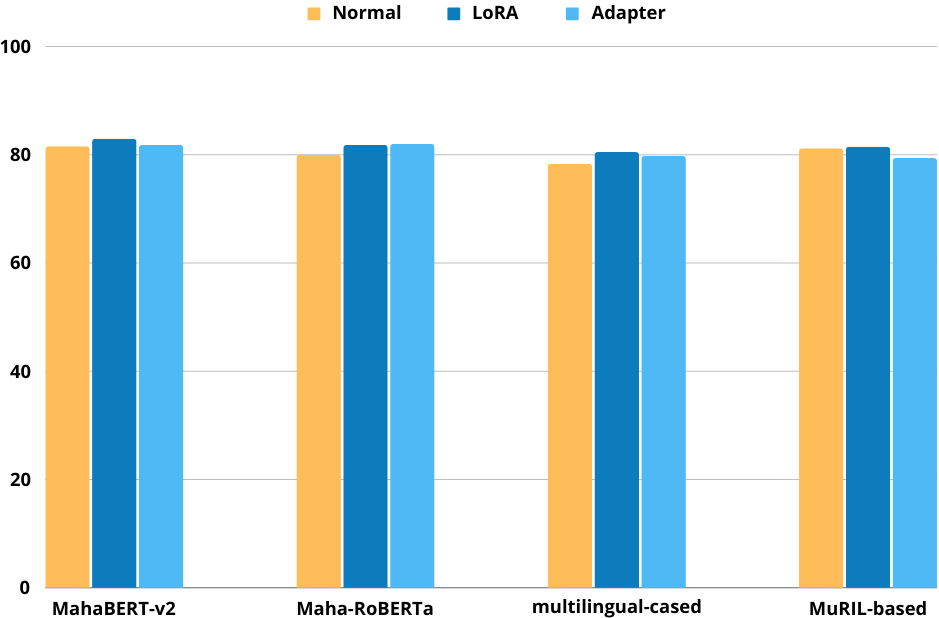}
    \caption{L3Cube MahaHate Accuracy}
    \label{fig:Datasets}
\end{figure}
\begin{table}[h]
    \centering
    \begin{tabular}{|c|c|c|c|}
        \hline
        \textbf{Models} & \textbf{Normal} & \textbf{LoRA} & \textbf{Adapter} \\
        \hline
        MahaBert-v2 &94	&94.19	&92.27 \\
        \hline
        Maha-Roberta & 90	&95	&93.11 \\
        \hline
        multilingual-cased & 93.29	&93.4	&94.94 \\
        \hline
        MuRIL-based &92.35	&94.27	&90.73 \\
        \hline
    \end{tabular}
    \caption{L3Cube MahaNews Accuracy Values.}
    \label{tab:Table IV}
\end{table}
\begin{figure}[!h]
    \centering
    \includegraphics[scale=0.20]{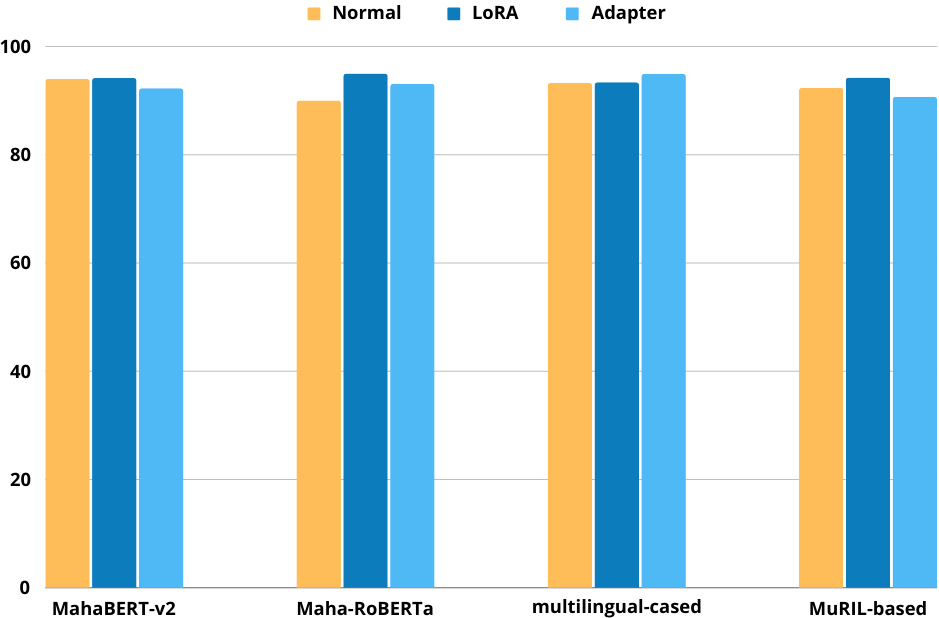}
    \caption{L3Cube MahaNews Accuracy}
    \label{fig:Datasets}
\end{figure}

\section{Result}\label{sec:Result}
Tables \ref{tab:Table II},\ref{tab:Table III} and \ref{tab:Table IV} show the results of the comparison of different language models on two natural language processing tasks: sentiment analysis and news classification. The models being compared are mahaBert-v2, maha-Roberta, multilingual-cased, and muril-based. The tasks are being performed on two different datasets: mahasent and mahaHate-4class for sentiment analysis, and mahaNews for news classification.

From the above tables, it can be inferred that in case of sentimental analysis MuRIL based model outperformed other models when trained without any PEFT parameters. It achieved accuracies of 84.93\% and 81.55\% with Mahasent and Mahahate respectively. It can also be observed that MahaBERT and MahaRoBERTa achieved better accuracies with LoRA and Adapter as the PEFT methodologies respectively. The Mahabert-LoRA combination achieved a score of 85.5\% for the Mahasent dataset and 82.9\% for the MahaHate dataset while the MahaHate-adapter methodology achieved scores of 84.09\% and 82\% for Mahasent and Mahahate datasets respectively.

For the news classification task MahaBert retains the maximum accuracy score of 94\% without using any PEFT methodologies. In case of LoRA and Adapter PEFT methodologies MahaRoberta and multilingual cased outperformed other models and achieved accuracies of 95\% and 94.94\% respectively.

Overall, it was seen that using PEFT methods along with the models comparative results can be achieved by reducing the training time and training parameters to a significant extent, eventually reducing the cost of training the model.

\section{Conclusion}\label{sec:Conclusion}
In this study we explore Marathi and, by extension, other Indic languages within the context of the emerging field of Natural Language Processing (NLP).Our paper offers thorough examination of mahaBert-v2,maha-Roberta,multilingual-cased,muril-based transformer models using accuracy as the main metric on well-known datasets like MahaSent, MahaHate, and MahaNews.

Furthermore, using these models, the paper presents and illustrates the use of Parameter Efficient Fine-Tuning (PEFT) techniques. This solves the important problem of computational costs in model development and greatly accelerates training speed without sacrificing accuracy. 

All in all, this study provides insightful information about how well the models work, which lays the groundwork for future improvements in natural language processing (NLP) for Marathi and related Indic languages. Researchers and practitioners looking to improve NLP applications in various linguistic contexts can benefit from the approaches and conclusions discussed in this paper as the digital content landscape continues to change.
\section*{Future Scope}
To further enhance the capabilities and impact of this project, several key areas can be targeted for future work, like enriching the training data with diverse domains like legal documents and social media dialects will broaden the model's real-world capabilities and adaptability. The usage of these PEFT methodologies can also be extended to other NLP tasks and deep learning methodologies. Additionally, exploring alternative PEFT techniques beyond those already used offers the potential for faster, more resource-efficient training, paving the way for faster model iterations and testing. Finally, evaluating on basis of other task-specific metrics like F1 score will provide a better picture of the model's strengths and weaknesses, guiding targeted improvements for specific NER applications.
\section*{Acknowledgements} 
This work was completed as part of the L3Cube Mentorship Program in Pune. We would like to convey our thankfulness to our L3Cube mentors for their ongoing support and inspiration.

\printbibliography
\end{document}